\documentclass[a4paper,12pt]{article}

\usepackage[ansinew]{inputenc}
\usepackage{natbib}
\usepackage{amsmath}    
\usepackage{amssymb}
\usepackage{graphicx}  

\def\intfunc{\mathop{\rm \sqcap}}
\def\heaviside{\mathop{\rm \Theta}}
\def\var{{\rm VAR}}
\def\bias{{\rm BIAS}}
\def\mse{{\rm MSE}}
\def\rmse{{\rm RMSE}}
\newtheorem{proof}{Proof}

\begin{document}

\title{\LARGE Generalized Prediction Intervals for Arbitrary Distributed High-Dimensional Data}
\date{}

\author{\small Steffen Kühn\\ \small Technische Universit\"at Berlin\\ \small Chair of Electronic Measurement and Diagnostic Technology\\ \small Einsteinufer 17, 10585 Berlin, Germany\\ \small steffen.kuehn@tu-berlin.de}

\maketitle

\begin{abstract}
This paper generalizes the traditional statistical concept of prediction intervals for arbitrary probability density functions in high-dimen\-sional feature spaces by introducing significance level distributions, which provides interval-independent probabilities for continuous random variables. The advantage of the transformation
of a probability density function into a significance level distribution is that it enables one-class classification or outlier detection in a direct manner.
\end{abstract}

\section{Introduction}

A prediction interval is an interval that will, with a specified degree of confidence, contain future realizations or, in the terminology of pattern recognition, feature vectors \citep{Hahn91}. The appeal of this concept is its clear stochastic meaning. The great disadvantage is that this definition is usually too restricted, for example for multimodal distributions. It is intuitively clear that, in this case, more than one interval for probable feature vectors can exist and it would be better to speak of prediction \textit{regions}. Even more complicated is the situation for high-dimensional feature spaces. This lack of generality is probably the reason why prediction intervals are rarely used in pattern recognition.

This is actually a pity, because prediction regions would be very useful, for example, for the recognition of outliers \citep{Barnett94} or the detection of novelty or normality. Instead of prediction intervals, numerous other methods are used for this purpose. They can be grouped roughly into two categories:
\begin{enumerate}
\item Distance-based and novelty or normality score-based approaches \citep{Knorr00, Dolia2002, Moonesignhe2006, Angiulli2006,  Ting2007},
\item Methods that introduce a separate rejection class in combination with a classifier \citep{Singh2004, Steinwart05}.
\end{enumerate}
If applying the method I propose here to outlier detection, it belongs to the first category with a probability as normality score. Before going into the details, I will give a short overview of related works.

Simple distance-based methods rely on the concept of the neighborhood of a point, for example, the $k$ nearest neighborhood \citep{Knorr00}. Outliers are those points for which there are less than $k$ points within a distance $\delta$ in the dataset. \cite{Ramaswamy00} propose a method to choose this threshold $\delta$ automatically based upon a dataset. The idea is to consider as outliers the set of points with the highest distances to their $k$th nearest neighbors. Of course, here is also a threshold necessary, but it has now a statistical reasoning as quartile of the $k$th nearest neighbor distance distribution, which simplifies the choice. A more recent article based on this idea is published by \cite{Angiulli2006}, who apply a weighted sum of the $k$th nearest distances per point. Although the idea is quite simple, the methods have low computation costs. Furthermore, they make only minor assumptions about the underlying distribution.

Another category of algorithms that are related to outlier detection is robust regression. The outlier detection is here more a means to an end, because the goal is to avoid that outliers influence the estimation of the regression function. This means that it is in this case sufficient to detect outliers indirectly. \cite{Ting2007}, for example, apply an outlier-score to control the influence that a point has in the parameter estimation process of the regression function. For this purpose, weights are introduced, which are estimated based on the assumption that the noise is Gaussian distributed. This is often sufficient, for example for most sensor signals. The algorithm is real-time capable, but far away from generality. This method also belongs to category one. 

The idea of the second category is very different. At the first glance, it seems to be impossible to use classifiers to detect outliers, because classifiers need for the estimation of their parameters samples from inliers \textit{and} outliers. Usually, only samples for inliers are available. The idea is to create a enclosing cloud of outlier samples synthetically with a random generator. Afterwards, it is possible to train the classifier. \cite{Singh2004}, for example, apply a neural network for this purpose. Other classifiers are also possible, for example, an SVM \citep{Steinwart05}. Regardless of the applied classifier, these probabilistic methods need for the generation of the hull a measure in which degree a generated sample point is an outlier. \cite{Singh2004}, for example, use for this purpose simple prediction intervals ($2.5\,\sigma$ ranges).   

In conclusion, both categories have to solve the same problem: to find an appropriate zero level set for the inlier generating density. In the subsequent sections I will show that this problem can be mapped to a choice of a significance level and that it is possible to generalize the traditional statistical concept of prediction intervals to prediction regions. 

\section{Prediction Regions}

A prediction interval denotes a region in which future feature vectors $\boldsymbol{x}$ occur with a predetermined probability. For its computation, it is essential that the generating probability density function $p_{\boldsymbol{X}}(\boldsymbol{x})$ for the random variable $\boldsymbol{X}$ is known. In this case 
\begin{equation}
P(\boldsymbol{x}_1 \leq \boldsymbol{X} \leq \boldsymbol{x}_2) = \int\limits_{\boldsymbol{x}_1}^{\boldsymbol{x}_2}\,p_{\boldsymbol{X}}(\boldsymbol{\omega})\,\mathrm{d}\boldsymbol{\omega}
\label{class_f30}
\end{equation}
is the probability for a future feature vector within the region $[\boldsymbol{x}_1,\boldsymbol{x}_2]$. 

For a prediction interval, the region borders have to be established so that the probability for outliers is lower than a given, fixed threshold -- the significance level $\alpha$. A typical example is $\alpha = 5\%$ and mean that the region is to be determined so that $95\%$ of all possible feature vectors fall into it. Usually, an infinite number of borders fulfill this requirement. Thus, for Gaussian distributions, the region is centered on the expectation value. But, this definition is only appropriate for this special case. 

The reason for this is the fact that the integration borders are defined directly. A way to solve this problem is to apply the level set idea with the probability density function $p_{\boldsymbol{X}}(\boldsymbol{x})$ as level set function. The set
\begin{equation}
\Gamma_{\theta} = \left\{\boldsymbol{x}\,|\,p_{\boldsymbol{X}}(\boldsymbol{x}) - \theta = 0\right\}
\end{equation}
can serve as implicit definition for the integration borders. The question is, how the threshold $\theta$ corresponds to the significance level $\alpha$ so that
\begin{equation}
\alpha \stackrel{!}{=} \int\limits_{p_{\boldsymbol{X}}(\boldsymbol{\omega}) \leq \theta} p_{\boldsymbol{X}}(\boldsymbol{\omega}) \mathrm{d}\boldsymbol{\omega}
\label{plau_f11}
\end{equation}
becomes true.

Let $F_Y$ be the cumulative distribution function of the probability density function values $y = p_{\boldsymbol{X}}(\boldsymbol{x})$ and $F_Y^{-1}$ its inverse. In this case, the threshold $\theta$ can be computed from a given significance level $\alpha$ by
\begin{equation}
\theta = F_Y^{-1}(\alpha).
\label{plau_f0}
\end{equation}
I now prove this assertion.\\ 

\begin{proof}
The feature vectors $\boldsymbol{x}$ can be interpreted as realizations of the \textit{vectorial} random variable $\boldsymbol{X}$. Because of this, $Y = p_{\boldsymbol{X}}(\boldsymbol{X})$ is also a random variable. But contrary to $\boldsymbol{X}$, $Y$ is \textit{scalar} valued. The relation between $\boldsymbol{X}$ und $Y$ is strictly deterministic, that is,
\begin{equation}
p_{Y|\boldsymbol{X}}(y|\boldsymbol{x}) = \delta(y - p_{\boldsymbol{X}}(\boldsymbol{x}))
\end{equation}
and consequently 
\begin{equation}
p_{Y,\boldsymbol{X}}(y,\boldsymbol{x}) = \delta(y - p_{\boldsymbol{X}}(\boldsymbol{x})) p_{\boldsymbol{X}}(\boldsymbol{x}).
\end{equation}
The marginalization in respect to $\boldsymbol{X}$ gives finally a formula to convert the probability density function $p_{\boldsymbol{X}}$ into the probability density function $p_{Y}$:
\begin{equation}
p_{Y}(y) = \int\limits_{\boldsymbol{\omega}} \delta(y - p_{\boldsymbol{X}}(\boldsymbol{\omega}))\,p_{\boldsymbol{X}}(\boldsymbol{\omega})\mathrm{d}\boldsymbol{\omega}.
\label{plau_f1}
\end{equation}
Now, we can calculate the cumulative distribution function $F_Y(y)$, which is defined by
\begin{equation}
F_Y(y) = \int\limits_{-\infty}^{y} p_{Y}(y')\mathrm{d}y'. 
\label{plau_f2}
\end{equation}
Inserting (\ref{plau_f1}) in (\ref{plau_f2}) delivers 
\begin{equation}
\begin{split}
F_Y(y) & = \int\limits_{-\infty}^{y} \int\limits_{\boldsymbol{\omega}} \delta(y' - p_{\boldsymbol{X}}(\boldsymbol{\omega}))\,p_{\boldsymbol{X}}(\boldsymbol{\omega})\mathrm{d}\boldsymbol{\omega}\, \mathrm{d}y'.
\label{plau_f3}
\end{split}
\end{equation}
With the Interval function 
\begin{equation}
\intfunc_a^b(x) = \left\{\begin{array}{l@{\quad}l} 1,& a\leq x \leq b \\ 0,&\mbox{otherwise}\end{array}\right.
\end{equation}
the expression (\ref{plau_f3}) can be transformed to
\begin{equation}
\begin{split}
F_Y(y) & = \int\limits_{\boldsymbol{\omega}} \intfunc_{-\infty}^{y}(p_{\boldsymbol{X}}(\boldsymbol{\omega}))\,p_{\boldsymbol{X}}(\boldsymbol{\omega})\mathrm{d}\boldsymbol{\omega} \\
& = \int\limits_{p_{\boldsymbol{X}}(\boldsymbol{\omega}) \leq y} p_{\boldsymbol{X}}(\boldsymbol{\omega})\mathrm{d}\boldsymbol{\omega}
\end{split}
\label{plau_f5}
\end{equation}
A comparison shows that the right side of expression~(\ref{plau_f11}) is identical to $F_Y(\theta)$. For this reason, we can write (\ref{plau_f11}) as
\begin{equation}
\alpha \stackrel{!}{=} F_Y(\theta).
\label{plau_f8} 
\end{equation}
The cumulative distribution function $F_Y$ is per definition monotonic. If it is even \textit{strictly} monotonic, the inverse function $F_Y^{-1}$ exists and we can compute $\theta$ for a given $\alpha$ by expression~(\ref{plau_f0}). Otherwise, we could obtain for one value of $\alpha$ an interval of possible values for $\theta$. Because all values in this interval result in the same $\alpha$ for the integration~(\ref{plau_f11}), any value in this interval can be used to solve the equation. 
\end{proof}

Summarization: A significance level $\alpha$, as it is usually applied in statistics, can be transformed with the \textit{scalar valued} function $F_Y^{-1}$ into a level set threshold $\theta$. With this threshold, it is possible to classify those feature vectors with a statistical significance of $1-\alpha$ as outlier, whose probability density function values $p_{\boldsymbol{X}}(\boldsymbol{x})$ are lower than $\theta$. The ranges for which the condition $p_{\boldsymbol{X}}(\boldsymbol{x}) \geq \theta$ is fulfilled are the prediction regions. 

\section{Significance Level Distributions}

It is shown that the cumulative distribution function $F_Y(y)$ can be used to transform a given significance level $\alpha$ into a level set threshold $\theta$. But, because of expression~(\ref{plau_f5}), $1-F_Y(y)$ can also be applied as measure to describe our degree of surprise about a certain feature vector $\boldsymbol{x}$. We summarize and define 
\begin{equation}
\fbox{$ \displaystyle
b_{\boldsymbol{X}}(\boldsymbol{x}) \; := \int\limits_{p_{\boldsymbol{X}}(\boldsymbol{\omega}) \leq p_{\boldsymbol{X}}(\boldsymbol{x})}^{\big.} p_{\boldsymbol{X}}(\boldsymbol{\omega})\mathrm{d}\boldsymbol{\omega} \; = \; F_Y(p_{\boldsymbol{X}}(\boldsymbol{x}))$ \quad}
\label{defsld}
\end{equation}
and name $b_{\boldsymbol{X}}(\boldsymbol{x})$ the \textit{significance level distribution} of $\boldsymbol{x}$ for the random variable $\boldsymbol{X}$.

The significance level distribution is in the true sense of the word a ``probability distribution'' because it provides a \textit{probability} (the significance level) for every \textit{continuous} realization $\boldsymbol{x}$. Unfortunately, the term ``probability distribution'' is already used for probability density functions, which do not provide probabilities but probability density values. Note that the significance level distribution does not deliver the probability for a single realization $\boldsymbol{x}$ itself, but the \textit{probability for all even more unlikely realizations than $\boldsymbol{x}$}. Nevertheless, $b_{\boldsymbol{X}}(\boldsymbol{x})$ provides valuable information for the assessment of the realization $\boldsymbol{x}$ and allows to decide if it is sure, probable, or only possible. 

\begin{figure}[th]
\centerline{\includegraphics[width=\columnwidth]{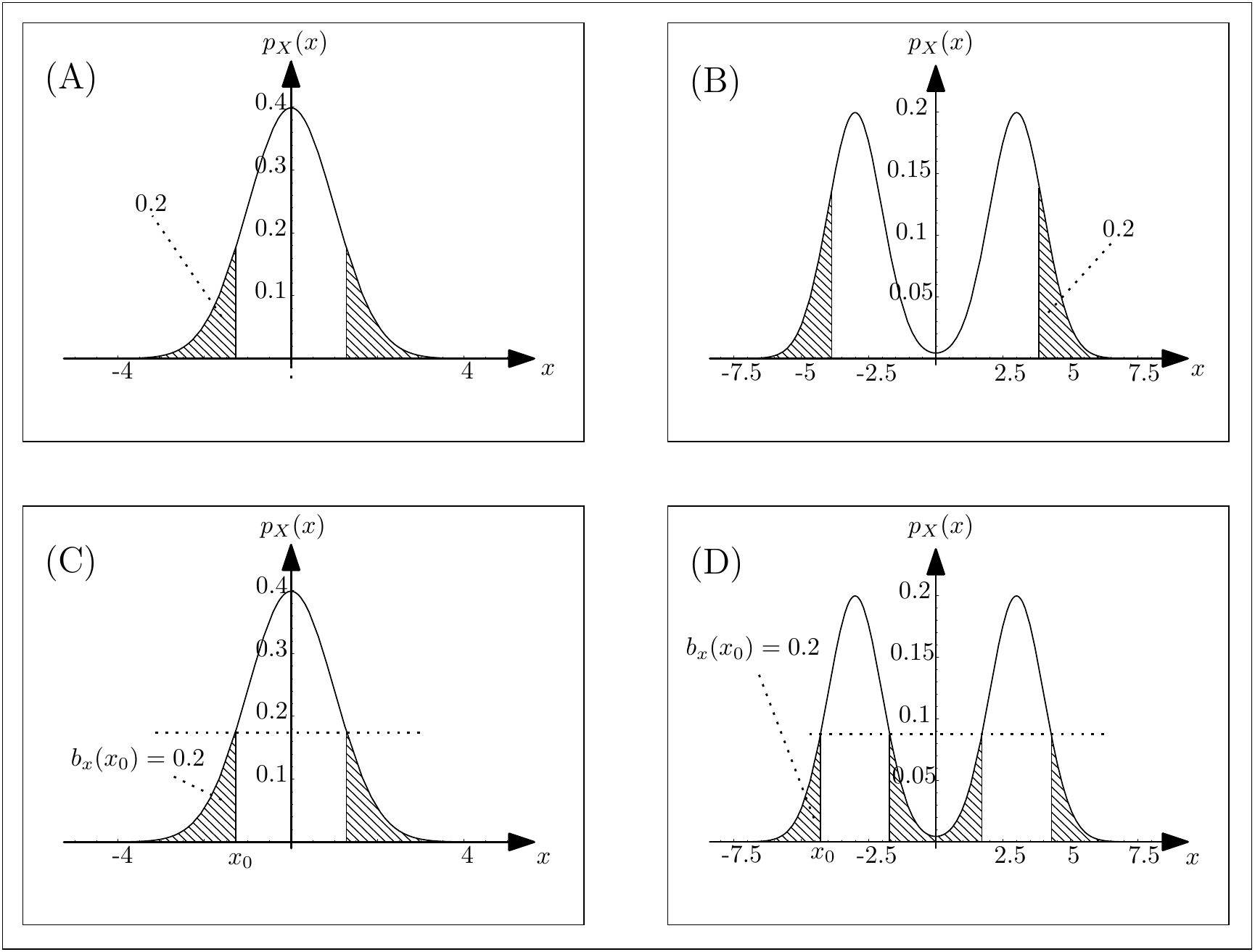}}
\caption{For unimodal and symmetric distributions the significance level distribution (C) and the prediction interval (A) lead to the same results. But for multimodal distributions only the significance level distribution is reasonable (D).}
\label{fig3}
\end{figure} 

\begin{figure}[th]
\centerline{\includegraphics[width=\columnwidth]{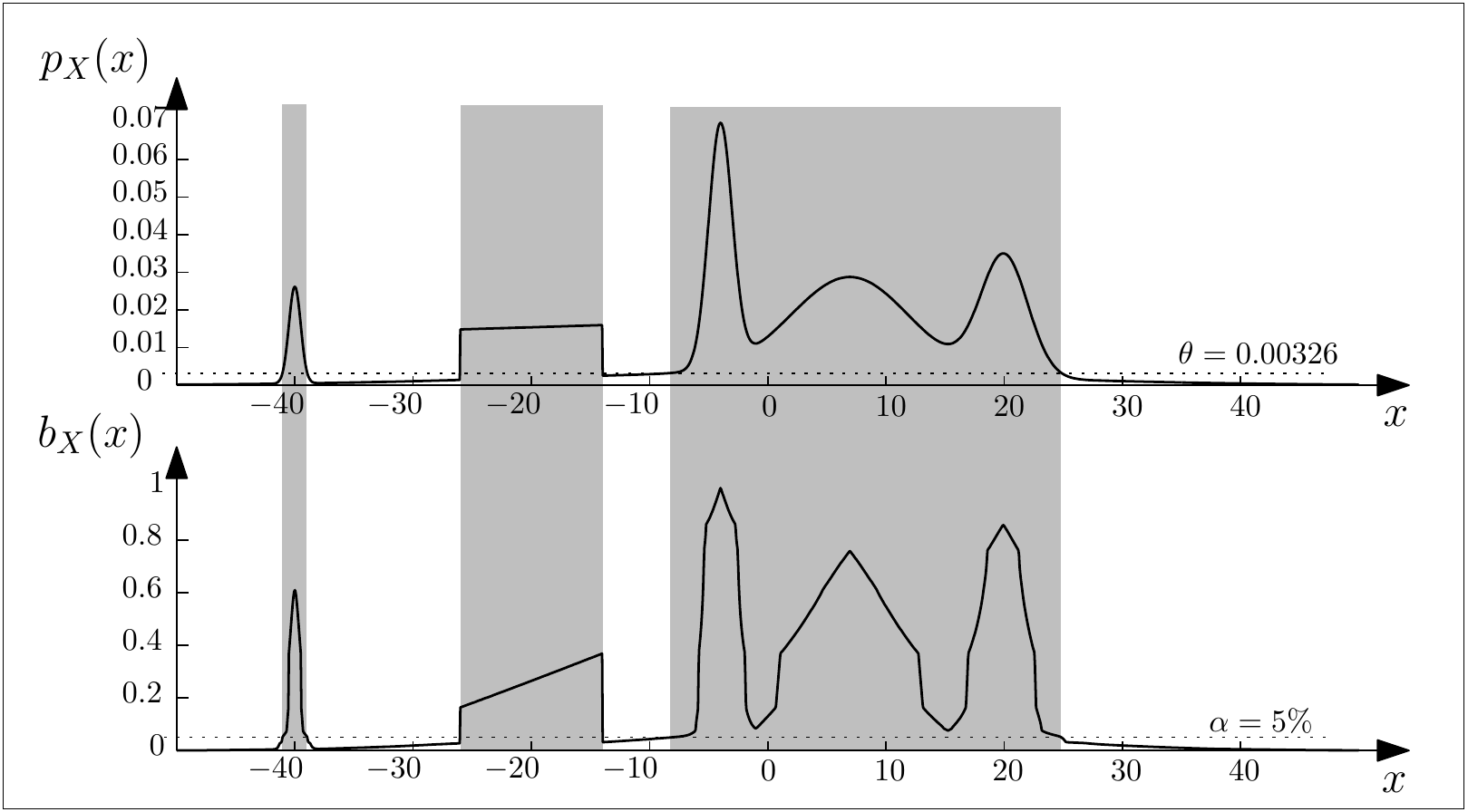}}
\caption{An example for a probability density function and its related significance level distribution. The white zones are the ``outlier regions'' for a significance level of $5\%$. The threshold $\theta = 0.00326$ corresponds to $\alpha = 5\%$.}
\label{class_fig2}
\end{figure}

For simple standard distributions, such as the Gaussian distribution or the Cauchy distribution, the significance level distribution can be given in closed form. Note that for a symmetric and unimodal distribution the significance level distribution and the prediction interval is identically (see Fig.~\ref{fig3}). For more complex distributions this is usually not valid and it is here seldom possible to give the significance level distribution in closed form. In these cases it is reasonable to estimate the cumulative distribution function $F_Y$. The next section~\ref{The Estimation of FY} proposes a method and investigates its convergence speed. Figure~\ref{class_fig2} shows an example of a significance level distribution for a non-trivial probability density function. Please note that significance level distributions are not restricted to the one-dimensional case.  

\section{The Estimation of $F_Y$ \label{The Estimation of FY}}

Contrary to the complicated integration~(\ref{plau_f11}), the expression~(\ref{plau_f8}) can be easily computed, if we estimate $F_Y$. We assume that the probability density distribution $p_{\boldsymbol{X}}(\boldsymbol{x})$ is known or can be appropriately estimated. With the knowledge of $p_{\boldsymbol{X}}(\boldsymbol{x})$, it is possible to generate $n$ correspondingly distributed random samples:
\begin{equation}
D_{\boldsymbol{X}} = \big\{\boldsymbol{x}_1,\ldots,\boldsymbol{x}_n\big\}.
\end{equation}
Now we can transform this dataset into a dataset of probability density function values
\begin{equation}
\begin{split}
D_{Y} &= \big\{p_{\boldsymbol{X}}(\boldsymbol{x}_1),\ldots,p_{\boldsymbol{X}}(\boldsymbol{x}_n)\big\} \\
& = \big\{y_1,\ldots,y_n\big\}.
\label{plaus_f25}
\end{split}
\end{equation}
With this dataset $D_Y$, the cumulative distribution function $F_Y$ can be estimated by
\begin{equation}
\tilde{F}_Y(y) = \frac{1}{n} \sum\limits_{k=1}^{n} \heaviside(y-y_k)
\label{plau_f7}
\end{equation}
with the Heaviside function 
\begin{equation}
\heaviside(y) = \left\{\begin{array}{l@{\quad}l} 1,& y \geq 0 \\ 0,&\mbox{otherwise}\end{array}\right..
\end{equation}
The Glivenko-Cantelli theorem guarantees the convergence of this empirical distribution function $\tilde{F}_Y(y)$ to the true cumulative distribution function $F_Y(y)$ for $n\to\infty$. Note that it is unnecessary to sum over all elements $y_i$ for the computation of expression~(\ref{plau_f7}) if the dataset $D_{Y}$ is sorted. In this case, a binary search with computation costs of $\Omega(\log(n))$ can be applied. 

It is possible to give the root mean squared error of this estimator in dependency on the size of the dataset $n$:
\begin{equation}
\rmse = \sqrt{F_Y\frac{1-F_Y}{n}} \leq \frac{1}{\sqrt{4\,n}}.
\label{plaus_f12}
\end{equation}
This formula makes it possible to calculate the number of samples $n$ necessary for a desired accuracy with a given significance level $\alpha = F_Y$. It is important that the convergence speed does neither depend on the generating density $p_{\boldsymbol{X}}(\boldsymbol{x})$ nor on the dimension of the original problem~(\ref{plau_f11}). It follows a proof of expression~(\ref{plaus_f12}).\\

\begin{proof}
We calculate the mean squared error $\mse$ by computing the expectation value 
\begin{equation}
\mathcal{E}\{(F_y - \tilde{F_y})^2\} = \int\limits_{-\infty}^{+\infty} \ldots \int\limits_{-\infty}^{+\infty} \,(F_y - \tilde{F_y})^2\, p_Y(y_1)\mathrm{d}y_1 \ldots p_Y(y_n)\mathrm{d}y_n
\end{equation}
of the squared error in respect to all elements in the dataset (\ref{plaus_f25}). The $\mse$ is usually written as sum 
\begin{equation}
\mse = \bias^2 + \var
\label{plaus_f16}
\end{equation}
with 
\begin{equation}
\bias = \mathcal{E}\{\tilde{F}_Y\} - F_Y
\label{plaus_f17}
\end{equation} 
and 
\begin{equation}
\var = \mathcal{E}\{\tilde{F}_Y^2\} - \mathcal{E}\{\tilde{F}_Y\}^2.
\label{plaus_f18}
\end{equation} 
For the estimation formula~(\ref{plau_f7}) is
\begin{equation}
\begin{split}
\mathcal{E}\{\tilde{F}_Y\} & = \frac{1}{n} \sum\limits_{k=1}^{n} \mathcal{E}\{ \heaviside(y-y_k) \} \\& = \frac{1}{n} \sum\limits_{k=1}^{n} \int\limits_{-\infty}^{+\infty}\ldots\int\limits_{-\infty}^{+\infty} \heaviside(y-y_k) p_Y(y_1)\ldots p_Y(y_n)\mathrm{d}y_1\ldots \mathrm{d}y_n
\\& = \frac{1}{n} \sum\limits_{k=1}^{n} \int\limits_{-\infty}^{+\infty}\heaviside(y-y_k) p_Y(y_k)\mathrm{d}y_k 
\\& = \frac{1}{n} \sum\limits_{k=1}^{n} \int\limits_{-\infty}^{y} p_Y(y_k)\mathrm{d}y_k = \frac{1}{n} \sum\limits_{k=1}^{n} F_Y = F_Y.
\label{plaus_f19}
\end{split}
\end{equation}
This shows that the estimator~(\ref{plau_f7}) is unbiased. Furthermore, we have
\begin{equation}
\begin{split}
\mathcal{E}\{\tilde{F}_Y^2\} \;=\; & \frac{1}{n^2} \sum\limits_{k=1}^{n} \sum\limits_{j=1}^{n} \mathcal{E}\{ \heaviside(y-y_k) \heaviside(y-y_j) \} \\ 
\;=\; & \frac{1}{n^2} \sum\limits_{k=1}^{n} \sum\limits_{j\neq k}^{n} \mathcal{E}\{ \heaviside(y-y_k)\} \mathcal{E}\{\heaviside(y-y_j) \} \\
& + \frac{1}{n^2} \sum\limits_{k=1}^{n} \mathcal{E}\{\heaviside(y-y_k)^2 \}.
\end{split}
\label{plaus_f20}
\end{equation}
Because of $\heaviside(y-y_k)^2 = \heaviside(y-y_k)$, we obtain
\begin{equation}
\begin{split}
\mathcal{E}\{\tilde{F}_Y^2\} & = \frac{1}{n^2} \sum\limits_{k=1}^{n} \sum\limits_{j\neq k}^{n} F_Y^2 + \frac{1}{n^2} \sum\limits_{k=1}^{n} F_Y \\
& = F_Y^2 - \frac{1}{n} F_Y^2 + \frac{1}{n} F_Y.
\label{plaus_f21}
\end{split}
\end{equation}
Inserting (\ref{plaus_f21}) and (\ref{plaus_f19}) into (\ref{plaus_f18}) yields 
\begin{equation}
\var = \mse = \frac{1}{n}\,F_Y\,\left(1 - F_Y\right).
\end{equation}
Finally, because of $\rmse = \sqrt{\mse}$, we obtain expression~(\ref{plaus_f12}). 
\end{proof}

\section{Overview of the method}

\subsection*{Estimation}

\begin{enumerate}
\item If the inlier generating density $p_{\boldsymbol{X}}(\boldsymbol{x})$ is unknown, estimate it with a suitable algorithm.
\item Choose a upper bound for the $\rmse$ and calculate the necessary number $n$ of random samples to generate: $n = 1/(2\,\rmse)^2$.
\item Generate the random samples $D_{\boldsymbol{X}}=\{\boldsymbol{x}_1,\ldots,\boldsymbol{x}_n\}$.
\item Compute the derived dataset $D_Y=p_{\boldsymbol{X}}(D_{\boldsymbol{X}})$.
\item Sort $D_Y$.
\end{enumerate}

\subsection*{Application}

\begin{enumerate}
\item Choose a significance level $\alpha$.
\item Compute the density value $y = p_{\boldsymbol{X}}(\boldsymbol{x})$ for the interesting feature vector $\boldsymbol{x}$.
\item Calculate the significance level value $z = b_{\boldsymbol{X}}(\boldsymbol{x})$ by computing $z = F_Y(y)$.
\item Classify $\boldsymbol{x}$ as outlier, if $z < \alpha$.
\end{enumerate}

\section{Experimental Validation of Equation (\ref{plaus_f12})}

Usually, the inlier generating density $p_{\boldsymbol{X}}(\boldsymbol{x})$ is unknown and has to be estimated. For this reason, the quality of the proposed method for the outlier detection depends significantly on the quality of the applied density estimation algorithm. The influence of the $F_Y$ estimation is, however, marginal, because the $\rmse$~(\ref{plaus_f12}) can be reduced arbitrarily -- in contrast to the estimation of $p_{\boldsymbol{X}}(\boldsymbol{x})$ -- by increasing the random sample number $n$. The following experiment verifies this by comparing a theoretically determined significance level distribution $b_{\boldsymbol{X}}$ to an estimated version $\tilde{b}_{\boldsymbol{X}}$.

\begin{figure}[th]
\centerline{\includegraphics[width=\columnwidth]{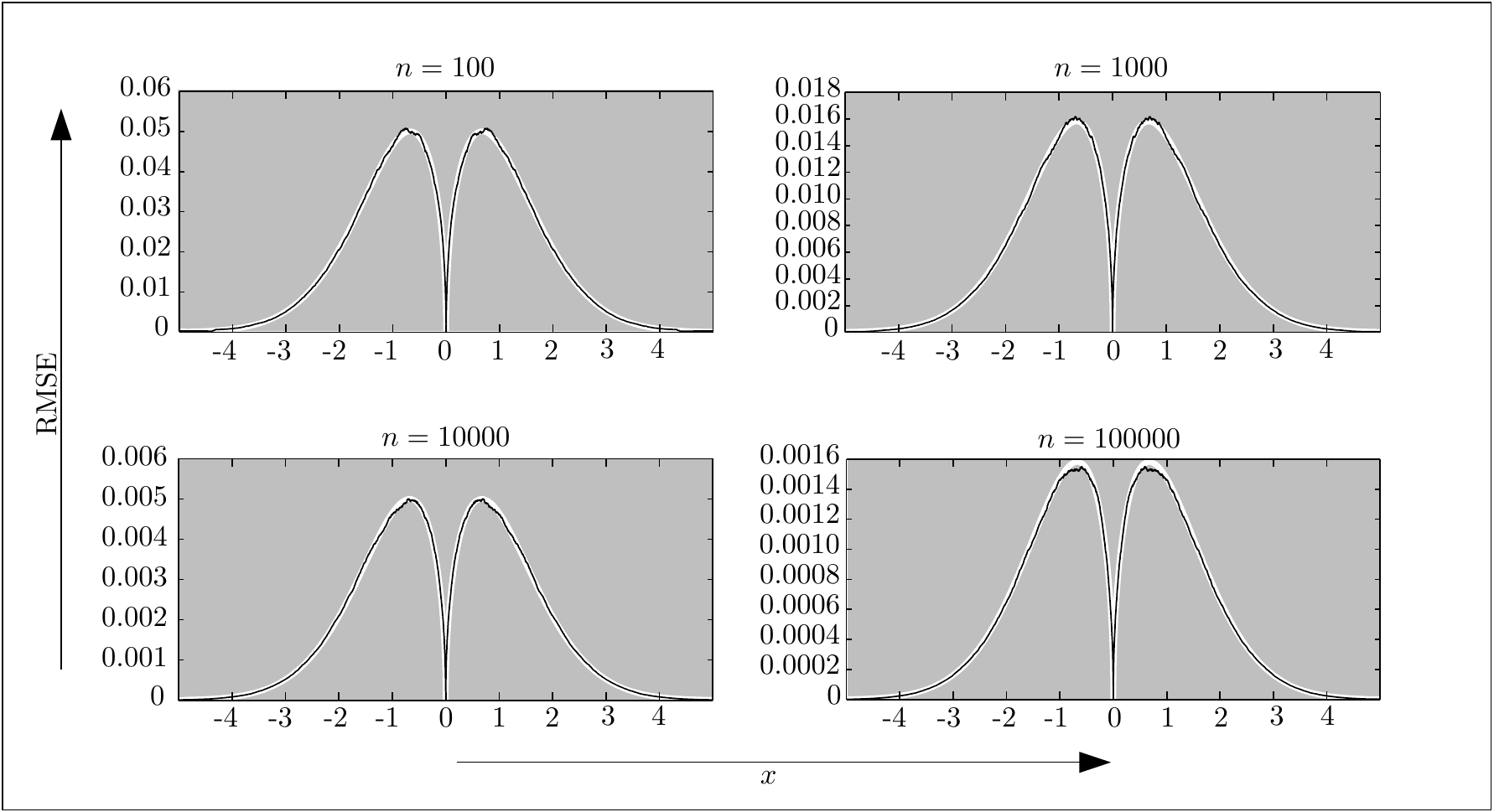}}
\caption{The figure shows the averaged root mean squared errors of 2000 single experiments for the significance level distribution estimation of a standard normal distribution by different random sample numbers $n$ (black lines) in comparison to the theoretical errors (fat white lines in the background).}
\label{class_fig1}
\end{figure}

It is only for some simple distributions possible to give the significance level distribution in closed form. Such an example is the Gaussian distribution
\begin{equation}
p_X(x) = \frac{1}{\sqrt{2\,\pi}\,\sigma} e^{-\frac{x^2}{2\,\sigma^2}}.
\end{equation}
The significance level distribution has, in this case, the form
\begin{equation}
b_X(x) = 1 - \mathrm{erf}\left(\frac{|x|}{\sqrt{2}\,\sigma}\right). \label{plaus_f30}
\end{equation}
In an experiment, I have estimated the significance level distribution for a standard normal distribution with the proposed method by varying the random sample number $n$. I have averaged for each value of $n$ the squared differences between the closed form~(\ref{plaus_f30}) and the estimated versions over 2000 single estimations. The results are summarized in Fig.~\ref{class_fig1}, which shows that the experiment confirms the theoretical predictions. 

\section{Conclusions}

In this article, I have shown that it is always possible to compute prediction regions as generalization of prediction intervals, no matter if the generating density is high-dimensional or multimodal. Only the density has to be known or estimated. 

The idea was to define the integration borders indirectly by a zero level set with the probability density function as level set function. This has lead to the problem of transforming the significance level defining a prediction interval into a level set threshold. I have shown that this can be easily accomplished by the cumulative distribution function of the probability density function values. The advantage is that the complicated integration in the high-dimensional feature space is mapped to a one dimensional function evaluation. 

Furthermore, I have introduced a new probability measure, the significance level distribution, which can be easily derived from the probability density function. The advantage is that it enables the assessment of the ``plausibility'' of an realization or feature vector because it provides probabilities also for continuous realizations. The transformation procedure has low computation costs and the estimation error of the method is negligible.

Please note that in practice the performance of the proposed method for one-class classification tasks depends  significantly on the quality of the applied density estimation method, just like the quality of a Bayes classifier for multi-class classification. On the contrary, for an optimal estimated density, the method would be necessarily optimal for one-class classification, just like a Bayes classifier is optimal for the multi-class classification case. Because density estimation it not the topic of this article I have deliberately omitted some experimental comparisions with other outlier recognition or one-class classification methods.


\end{document}